# Ethical Considerations When Constructing Participatory Design Protocols for Social Robots

Douglas Zytko
Computer Science and Engineering
Oakland University
Rochester Hills, Michigan, USA
zytko@oakland.edu

Wing-Yue Geoffrey Louie
Electrical and Computer Engineering
Oakland University
Rochester Hills, Michigan, USA
louie@oakland.edu

## ABSTRACT

Participatory design has emerged as a popular approach to foreground ethical considerations in social robots by incorporating anticipated users and stakeholders as designers. Here we draw attention to the ethics of participatory design as a method, distinct from the ethical considerations of the social robot being co-designed. More specifically, we consider the ethical concerns posed by the act of stakeholder participation - the morals and values that should be explicitly considered when we, as researchers or practitioners, devise protocols for participatory design of social robots ("how" stakeholders participate). We use the case of robot-assisted sexual violence mitigation to exemplify ethical considerations of participatory design protocols such as risk of harm, exploitation, and reduction of stakeholder agency. To incorporate these and other ethical considerations in the creation of social robot participatory design protocols, we advocate letting stakeholders design their own form of participation by including them in the creation of participatory design sessions, structures, and processes.

## KEYWORDS

Social Robots, Participatory Design, Ethics

## 1 Introduction

Ethics have become a core consideration in human-robot interaction (HRI), to the point of having its own term: roboethics, which broadly envisions robot design and development being guided by the intent to augment human society while preventing "misuse against humankind" [9]. Roboethics comprise the ethical considerations and problems that arise during interactions between human and robot, which are particularly germane to social robots given that their purpose centers on social interaction with humans, thus posing ample opportunity for both positive and (inadvertently) negative impact on humans. Beyond classic examples of moral dilemmas involving death, we can take cues from other emerging technologies like AI to foresee ethical concerns that are on the horizon for social robots such as racial and gender inequality and reinforcement of unhealthy behavioral patterns.

Ideally, ethical considerations of a social robot can be identified and directly designed for - what STS scholars have described as "materializing morality" [8]. The literature across HRI and HCI has elucidated different ways to explicitly incorporate ethics into social robot design. One is value sensitive design [3], which comprises a series of methods to provide designers with a comprehensive understanding of human values and morals held by anticipated users or the general public that can underpin design (see another example with AI, a similarly emerging technology, in [10]).

Another approach is participatory design, which at a fundamental level entails the involvement of anticipated users and other stakeholders into the design process *as* designers [7]. They work alongside professional designers and developers and share in decision-making to ensure that their goals, problems, and perspectives are reflected in the end product. Participatory design can be a powerful approach to designing ethical human-robot interactions because stakeholders are empowered to provide more than the morals and values that they would want foregrounded in social robots by other (professional) developers. They can also propose and iterate on their own designs for enacting those morals and values. For example, Axelsson and colleagues [2] provide a series of "canvases" to scaffold participatory design for social robots, one of which centers specifically on ethics of the robot. The canvas prompts co-designers to reflect on six different ethical problems that the robot could potentially contribute to such as physical safety risk, inequality, and reinforcement of inappropriate behavior, followed by prompts for solutions to such concerns.

In this workshop paper we draw attention to the ethics of participatory design as a method, distinct from the ethical considerations of the social robot being co-designed. More specifically, we consider the ethical concerns posed by the act of stakeholder participation - the morals and values that should be explicitly considered when we, as researchers or practitioners, devise protocols for participatory design of social robots ("how" stakeholders participate). There is a wide range of ethical considerations that apply to participatory design of social robots due to the extensive timelines needed for their design and development, the variety of expertise areas that could or should be incorporated in design, and the fluctuating levels of social robot familiarity that stakeholders may bring into design exercises.

Three ethical considerations for social robot participatory design protocols that we highlight here include:

- **Harm:** How may the act of design participation expose a stakeholder to harm? We conceptualize harm through participation in various ways, which could manifest physically, emotionally, socially, cognitively, and so on. Some examples include risk of physical harm through emergent robot behaviors and emotional harm through the forced reliving of traumatic experiences that serve as the foundation of a stakeholder's designs for the robot, which may become more vivid as the robot is materialized. How do we craft participatory design activities and structures that are sensitive and mitigative to harm?

- **Exploitation:** Participatory design may be within the job responsibilities of the practitioners and researchers conducting such processes, thus providing inherent professional (and by extension, financial) benefit to them. Are the benefits to stakeholders contributing to robot design similarly inherent or guaranteed? How do we recognize and ensure the truly anticipated benefits of stakeholder participation, especially given the relatively long timeline for social robot development and evaluation?

- **Agency:** Participatory design protocols, especially in HRI, are often intricately structured and supervised through predetermined activities, boundaries, and rules. While these structures can remedy stakeholder confusion and maintain stakeholder focus on the task at hand, they may also usurp stakeholder agency - their freedom over how they participate. Are stakeholders being supported and empowered to participate in the manner that they want?

To incorporate these and other ethical considerations in the creation of social robot participatory design protocols, we advocate expanding the boundaries of participatory design to also include the protocol for participation. In other words, we advocate including anticipated stakeholders in the creation of participatory design sessions, structures, and processes that they—or those aligning with their particular demographic—may eventually design within. For the rest of this workshop paper, we use one of our ongoing projects about robot-assisted sexual violence mitigation to exemplify ethical considerations of participatory design protocols and approaches to involving stakeholders in devising their own mode of participation.

## 2 The Case of Robot-Assisted Sexual Violence Mitigation

Sexual violence involves any "sexual act that is committed or attempted by another person without freely given consent of the victim" [10]. Examples include rape (unwanted vaginal, oral, or anal penetration) and unwanted physical sexual contact (e.g., kissing, touching of the body) [91]. The field of human-computer interaction (HCI) has studied and proposed various technology-facilitated tools for sexual wellness as well as solutions to sexual violence. To the latter, examples include wearable devices that emit bad odors to deter rapists [5], panic buttons on mobile devices to alert authorities [1], AI to detect sexual predators who attempt to coerce minors into meeting face-to-face [6] and dating apps transformed into sexual consent apps [4]. However, these solutions are typically either known to be unsuccessful, severely limited, or are lacking empirical evidence of their impact.

Social robots are an opportune, alternative technology to consider for prevention of sexual violence because they can be physically present and actively provide assistance when technology mediation is most needed: when nonconsensual sex is about to occur. To clarify, we imagine social robots serving roles beyond physical protection against perpetrators intent on causing sexual harm – we imagine social robots more as mediators of sexual interactions that ensure mutual exchange of consent to sexual activity. There are endless directions for this broad vision of robot-assisted sexual experience.

Participatory design is an appropriate method for pursuing robot-assisted sexual violence mitigation because the success of such robots is contingent on willingness of users to incorporate them into their lives—especially intimate contexts—and perceive such robots as normal and acceptable additions to their sexual experiences.

Our team has confronted significant ethical considerations in planning a participatory design process for such a social robot, which we summarize around the three categories listed in the previous section:

- **Harm – risk of sexual violence re-traumatization:** Sexual violence is a traumatic experience for survivors, and situations in which they are forced to remember and even describe their sexual violence experiences to others can incur re-traumatization and severe emotional distress. There are also risks of physical harm as co-design progresses towards tangible artifacts and materials. How can we empower stakeholders to participate in designing robot-assisted sexual violence prevention solutions while managing risks of harm?

- **Exploitation – the mounting costs of designing sexual violence prevention robots:** Social robot design is typically a time-consuming process, meaning that the chances of harm and re-traumatization are elevated as a given stakeholder chooses to sustain participation. Stakeholders, due to their relative unfamiliarity with participatory design and social robots, may not be able to predict these mounting emotional costs of participation up front, which may result in benefits of participation eventually failing to supersede such costs. How can we as researchers protect stakeholders from inadvertent exploitation in the face of uncertain costs of participation in social robot co-design?

- **Agency – how should consensual sex occur?:** One reason sexual violence persists is because of varying conceptions of sexual consent, or the ways in which a person gives and (perceives to) receive agreement to a sexual activity [11]. There is risk of usurping stakeholder agency if the robot participatory design process imposes a particular consent exchange process onto stakeholders that is "desirable" to researchers, but contrary to how stakeholders may envision consensual sex occurring. How can we structure participatory design activities so that stakeholders can productively contribute to design of social robots for sexual experiences without imposing preconceived solutions and values onto them?

To address these ethical considerations, we are engaging in a series of informal "protocol construction" sessions with stakeholders who represent demographics at heightened risk of sexual violence (e.g., women, members of the LGBTQIA+ community), practitioners of sexual violence victim care, and researchers of sexual violence perpetration. The intent of these conversations is to produce a participatory design study protocol with the given stakeholder that assuages their concerns about the aforementioned categories of harm, exploitation, and agency. While the specifics of these conversations are contingent on the stakeholder's prior knowledge, we have a pre-prepared slide deck for key concepts such as participatory design and HRI that we employ to assist the stakeholder in brainstorming. Upon completion of these sessions, we intend to qualitatively analyze the stakeholder-produced study protocols to identify themes and patterns and ultimately produce a singular study protocol that accommodate the ethical considerations of the stakeholders. We look forward to sharing our experiences in the workshop and participating in discussion about the ethics of participatory design for social robots.

## 3  Author Bios

**Douglas Zytko** received his PhD in Information Systems from New Jersey Institute of Technology. He is currently an Assistant Professor in the Department of Computer Science and Engineering at Oakland University. He is also Director of the Oakland HCI Lab, a hub for interdisciplinary research into online-to-offline harm. The lab integrates researchers in human-computer interaction, human-robot interaction, AI, VR, psychology, and nursing to leverage emerging technologies for the prevention of harms that emerge through the combination of computer-mediated and face-to-face interaction. Doug is currently using participatory design methods for the prevention of online-to-offline sexual violence. The work puts stakeholders at risk of sexual violence victimization in position to articulate new use cases for emerging technologies such as social robots and AI pursuant to safety. The research also coalesces sexual violence experts in clinical and research contexts to produce theory-informed technologies for altering behavior of potential perpetrators.

**Wing-Yue Geoffrey Louie** received his B.Sc and Ph.D degrees in Mechanical Engineering from the University of Toronto (Canada) in 2011 and 2017, respectively. During his Ph.D, he received the Natural Sciences and Engineering Research Council of Canada Postgraduate Scholarship to develop and investigate approaches to personalize social robots healthcare applications. In 2017, he joined the Department of Electrical and Computer Engineering as an Assistant Professor at Oakland University (USA). He is presently the Director of the Intelligent Robotics Laboratory where he is now the principal investigator for the National Science Foundation project on developing approaches to enable healthcare professionals to teach social robots communication strategies for effective intervention delivery and University of Michigan's Automotive Research Centre project on studying human-robot interactions in video game engines. The core theme of his research is on the development of robot technology that can be easily customized by non-experts and personalized according to their needs. His research has been integrated in social and service robots for applications including therapy for autism spectrum disorder, older adult care, physiotherapy, early childhood education, search and rescue, and autonomous driving.